\def\year{2019}\relax
\documentclass[letterpaper]{article} %DO NOT CHANGE THIS
\usepackage{aaai19}  %Required
\usepackage{times}  %Required
\usepackage{helvet}  %Required
\usepackage{courier}  %Required
\usepackage{hyperref}
\hypersetup{
    colorlinks=true,
    linkcolor=blue,
    filecolor=magenta,      
    urlcolor=cyan,
}

\usepackage{graphicx}  %Required]
\usepackage{multirow}
\usepackage{listings}
\usepackage{colortbl}

\nocopyright

\lstdefinelanguage{PDDL-planimation}
{
  keywordstyle=\color{magenta},
  basicstyle={\small\ttfamily},
  sensitive=false,    % not case-sensitive
  morecomment=[l]{;}, % line comment
  alsoletter={:,-},   % consider extra characters
  morekeywords={
    define,domain,problem,not,and,or,when,forall,exists,either,
    :domain,:requirements,:types,:objects,:constants,
    :predicates,:action,:parameters,:precondition,:effect,
    :fluents,:primary-effect,:side-effect,:init,:goal,
    :strips,:adl,:equality,:typing,:conditional-effects,
    :negative-preconditions,:disjunctive-preconditions,
    :existential-preconditions,:universal-preconditions,:quantified-preconditions,
    :functions,assign,increase,decrease,scale-up,scale-down,
    :metric,minimize,maximize,
    :durative-actions,:duration-inequalities,:continuous-effects,
    :durative-action,:duration,:condition,
    :visual,:type,:properties,prefabImage,showName,x,y,color,width, height,prefabImage,depth,label,:image,equal,assign,:effects,
    distributex,distributey,distribute_within_objects_vertical,distribute_within_objects_horizontal,distribute_grid_around_point,settings,calculate_label,align_middle,apply_smaller,draw_line
  }
}

\newcommand{\pddlfont}{\lstinline[language=PDDL-planimation]}

\begin{document}
	\title{Planimation}
\author{	
	 Gang Chen,
	 Yi Ding,
	 Hugo Edwards,
	 Chong Hin Chau, 
	 Sai Hou,
	 Grace Johnson,
\\{\bf \Large
	 Mohammed Sharukh Syed,
	 Haoyuan Tang,
	 Yue Wu,  
	 Ye Yan,
	 Gil Tidhar
	 and Nir Lipovetzky}\\ \normalsize  
	 The University  of Melbourne \\ \normalsize 
	 Melbourne, Australia \\ \small nir.lipovetzky@unimelb.edu.au 
}

	\maketitle
	
	\begin{abstract}
	Planimation is a modular and extensible open source framework to visualise sequential solutions of planning problems specified in PDDL. We introduce a preliminary declarative PDDL-like animation profile specification, expressive enough to synthesise animations of arbitrary initial states and goals of a benchmark with just a single profile.\footnote{Best ICAPS 19 - Systen Demo Award - technical report}
	\end{abstract}
	
%	{\centering
%		Keywords: PDDL, Planning Visualization
%	}

\section{Introduction}

The adoption of a standard declarative specification of planning tasks through PDDL \cite{mcdermott:pddl-spec}, fostered by International Planning Competitions \cite{mcdermott:pddl}, has boosted the development of solvers by the research community. PDDL specifies a model of the problem, and the planner synthesises a solution which can take the form of a sequence, policy or a tree, depending on the model being solved \cite{geffner:book}. Solutions for classical planning take the form of a sequential plan expressed as text-based keywords identifying the \emph{grounded} actions mapping the initial state to a goal state. Special tools exist to validate the soundness of a plan \cite{howey:val} given a problem specification in PDDL, and assist on the detection of errors in existing solvers. Other tools have been developed to assist the PDDL modeling process \cite{vaquero:itsimple,barreiro:europa}. For instance, itSIMPLE assists the analysis of PDDL by translating state chart diagrams encoded by a modeler into Petri-Nets \cite{vaquero:itsimple}. Other tools provide insights about \emph{solvers} through search tree visualisations of a given search algorithm \cite{magnaguagno:webplanner}, or visualisations of the internal decision making process of a solver \cite{chakraborti:visualizations}. 

If the PDDL specification is syntactically correct but fails to model the classical planning problem the modeler had in mind, then there is no tool to give feedback and facilitate detecting the source of failure. For example, missing preconditions are prevalent and hard to detect if the only feedback is the name of the actions in a valid plan. \emph{Planimation}, a plan visualizer, intends to close the feedback loop and animate a plan given a PDDL specification, relying on visual cues to help modelers find the sources of mental and model misalignments. Furthermore, the tool diminishes the effort needed to understand and explain the dynamics encoded by PDDL problems, as plans explain themselves visually.

PDDL encodes a transition system declaratively by providing typically a single \emph{domain} file specifying actions and predicates, and a \emph{problem} file specifying objects, the initial state and goal states. Actions changing the valuation of predicates specify the state transition we aim to animate visually. In order to do so, we provide a third PDDL-like \emph{animation} file that specifies a sprite for each object, and the animation behaviour triggered \emph{when a predicate becomes true in a state}. The declarative visual animation language decouples the visualisation engine in the same way  PDDL decouples models from solvers. PDDL modelers can extend their problems with a single animation profile and visualize the plans returned by existing solvers. 

\section{Animation Profile}

The visualization is rendered in a two dimensional canvas. The animation profile allows the specification of the following properties for each \pddlfont{object} or \pddlfont{type}:
\pddlfont{x} and \pddlfont{y} as integer coordinates or null value; \pddlfont{color} as a hexadecimal RGB value, pre-specified colour constant or random color; \pddlfont{width} and \pddlfont{height} of the object; \pddlfont{prefabImage} as a base64 string of the object's image\footnote{Base64 image generator: {\scriptsize \url{https://www.base64decode.org/}}.}; \pddlfont{depth} of the object in the canvas; a boolean flag \pddlfont{showname} to specify whether to display the object' name; and an optional \pddlfont{label} string to substitute the default name of the object in the canvas. Given that many domains do not specify all the objects in the problem, we also allow for \pddlfont{custom} object declaration, which  adds a visual object in our canvas  and permits any predicate to animate it. For example, Blocksworld does not specify objects for the claw and table, which can be added as custom objects.

Predicates defined in the domain file are used by the animation profile to specify the animation behaviour applied to either a specific object, a set of objects, or the objects referred in the predicate parameters. Behaviours are specified by changing any of the object's properties using the {\pddlfont{equal}} operator in the {\pddlfont{:effects}} section of the predicate. For example, the predicate {\pddlfont{(on ?b1 ?b2)}} triggers the effects {\pddlfont{(equal (?b1 x) (?b2 x))}} and {\pddlfont{(equal (?b1 y) (add (?b2 y) (?b2 height) 2))}}, changing the $x$ value of block {\pddlfont{?b1}} to the $x$ value of {\pddlfont{?b2}}, and the $y$ value of {\pddlfont{?b1}} to the $y$ value of {\pddlfont{?b2}} plus its $height$ and a slack of 2 pixels. Visual effects on  predicate objects are sufficient to animate state transitions, but they cannot capture fully the visual layout specified in the initial state. 

In Blockworld, the initial state contains several $ontable$ predicates which can easily specify the y coordinate of a block, but not the x value, as its horizontal position is relative to the number of blocks in the table. For this purpose, we developed special \emph{constraint functions} for visual layouts, inspired by the use of Cassowary toolkit \cite{badros:constraint,badros:cassowary} in commercial applications such as Mac OS X and IOS 6. The functions supported so far by Planimation are {\pddlfont{distributex}}, which changes the $x$ value of a set of objects and distributes them along a horizontal plane; {\pddlfont{distributey}} which changes the $y$ value instead and distributes a set of objects vertically;  {\pddlfont{distribute_within_objects_vertical}} and {\pddlfont{distribute_within_objects_horizontal}} distribute a set of objects within the bounded visual space of another object (e.g cities); and {\pddlfont{distribute_grid_around_point}} which distributes a set of objects in a grid like structure. These functions can override the $x$ and $y$ values of an object. A function is invoked using the {\pddlfont{assign}} keyword in the effects of a predicate, and changes properties of the objects specified in the parameters of the predicate and the function. When a function $f$ is invoked by a predicate $p$, instead of being applied only over the objects of a single predicate $p$, it is applied over all the objects of all the true instantiations of $p$ in a state $s$. That is, the set of objects $O$ of a given function $f$ in a predicate $p$ and state $s$ is defined as $O(f,p,s)=\{ o \ | \ o\in obj(f),\ f \in eff(p),\  p\in s\}$ , where $obj(f)$ specifies the objects in function $f$ that appear in predicate $p$. For example, the initial state of Blocksworld is visualized by setting the effect of {\pddlfont{(ontable ?b)}} as {\pddlfont{(assign (?b x) (function distributex (objects} \pddlfont{?b) (settings (spacebtwn 40))))}}. This function is  applied over all the blocks that appear in $ontable$ predicates true in the initial state, and distributes them along a horizontal line 40 pixels apart. A second effect {\pddlfont{(equal (?b y) 0)}} assigns a block to the default $y$ position of the table. The full animation profile for Bloscksworld can be viewed at \url{tinyurl.com/yxlt96fp}.
% Read Only session: http://editor.planning.domains/#read_session=yiCWKZREGv
% Read/Write session: http://editor.planning.domains/#edit_session=f2FOIWszKeALAXO
Other special purpose visual functions are {\pddlfont{calculate_label}}, which returns the cardinality of a set of objects (e.g number of packages in a truck); {\pddlfont{align_middle}}, which aligns an object in the center of another object; {\pddlfont{apply_smaller}}, which changes the width or length of an object; and {\pddlfont{draw_line}}, which draws a line between two objects.

All these functions have been developed taking into account the diversity of IPC domains. More functions may be needed in the future, but these have been sufficient to animate a variety of domains, namely \emph{Blocksworld, Grid, Logistics} and \emph{Towers of Hanoi}. All animations are translations in space, scaling, or appear/disappear effects on objects.
For more information we refer the reader to the documentation in \url{planimation.github.io/documentation/}. %animation profile examples 

%% use pddl package for highlight

\section{Planimation}
Planimation is a modular and extensible open source framework to visualise sequential solutions of planning problems specified in PDDL. The framework has two main components, the Unity WebGL \emph{visualiser engine}, and a web API \emph{visualiser solver} written in Python. The solver can either load a saved plan visualisation, or interact with a planner through the API of \url{solver.planning.domains} \cite{muise2020keps}. Given a domain, problem and animation file, the visualisation solver invokes a planner, retrieves the solution and generates the animations. A visualisation file generator (VFG) can serialise the animation into a json file. Code and documentation for extensions and deployment on dedicated servers are included in \url{github.com/planimation}. Planimation runs in the web browser and is accesible through \url{planimation.planning.domains}.

\subsection{Functionality}

\begin{figure}[t!]
	\centering
	\includegraphics[height=5.5cm]{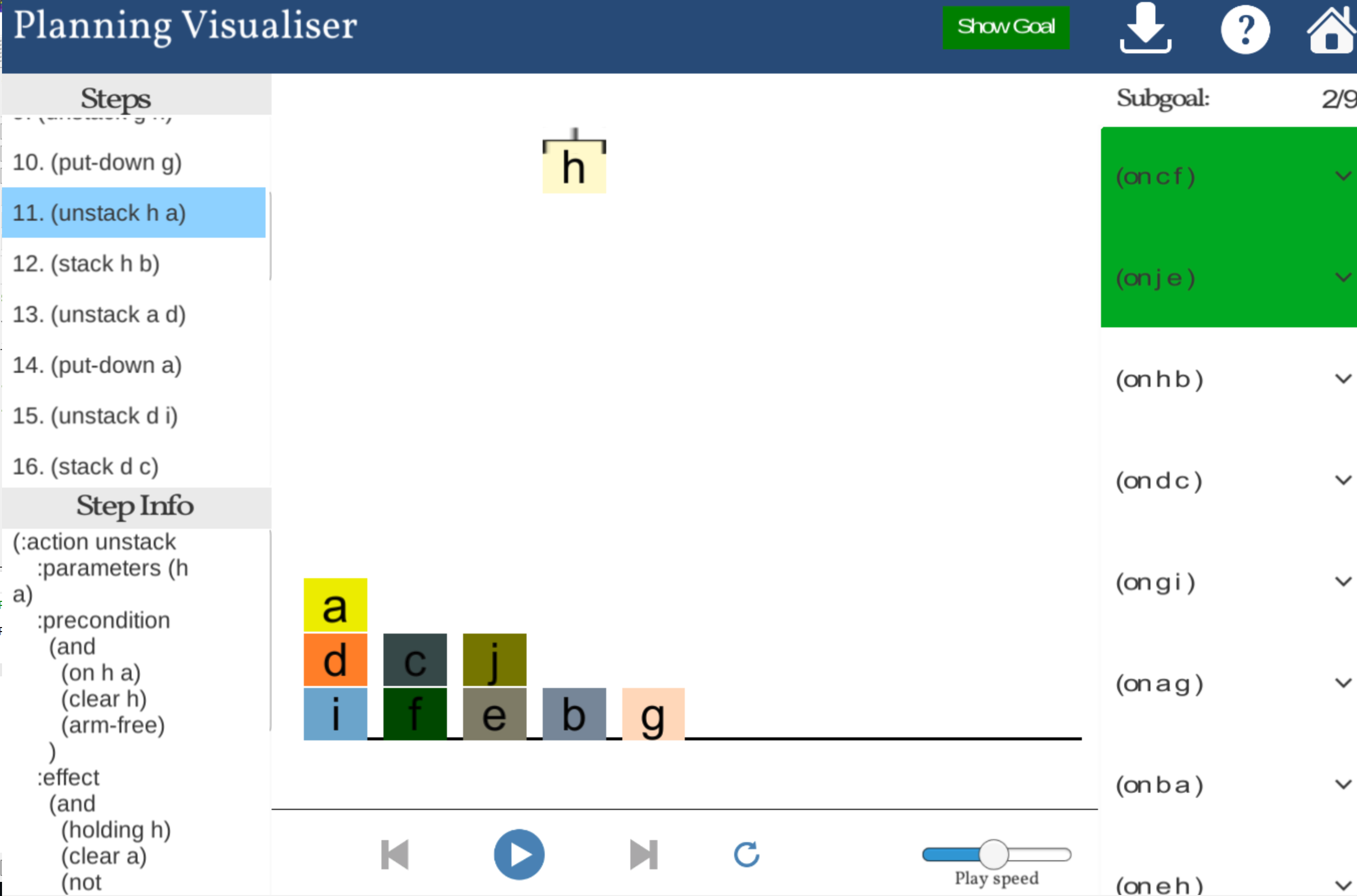}
	\caption{Planimation animating Blocksworld}
	\label{fig:blocks}
	\vspace{-0.2cm}
\end{figure}

Planimation UI is shown in Figure~\ref{fig:blocks}. It consists of a \emph{steps panel}, which shows the plan and action currently being executed. Any action $a$ is clickable, and sets the visualisation to the resulting state $s'= f(s,a)$. The \emph{Step Info} panel shows the preconditions and effects of the action being executed;  the \emph{Animation} panel displays the animation canvas and \emph{darkens} objects if they are in their goal position; the \emph{Control} panel allows animation speed changhes and playback controls;  the \emph{Subgoal} panel shows all the goal predicates and changes their colour when they are satisfied. Clicking on a subgoal opens a dropdown list with all the steps in which the subgoal is satisfied, jumping to such step if selected. The last panel is the \emph{header}, which contains help information, a button to show the goal state, and a button to download the visualisation in a json VFG format, GIF or WebM video. The VFG file can be used to load planimation without invoking a planner.

\section{Conclusion}
%``I always thought something was fundamentally wrong with the universe'' \citep{}
Planimation is intended to make solutions more amenable to humans interacting with planners, and assist in the modeling process as well as in the education of AI planning. Such plan visualisations can be part of an explainable AI planning solver \cite{fox:explainable} interacting with humans. Planimation has been developed by students as a final year software project, with the vision that the planning community continues its development. 

\subsubsection{Acknowledgments}

Naser Soueid and Guangling Yang for preliminary implementations, Gil Tidhar for the supervision of the team, and Christian Muise, Guillem Frances, and Miquel Ramirez for giving feedback on early versions.
\bibliographystyle{aaai}
\bibliography{control}
\end{document}